\documentclass[letterpaper, 10 pt, conference]{ieeeconf}  

\IEEEoverridecommandlockouts                              

\usepackage{graphicx} 
\usepackage{float}
\usepackage{mathptmx} 
\usepackage{amsmath} 
\usepackage{amssymb}  
\usepackage{multirow}
\usepackage{booktabs}
\usepackage{caption}
\usepackage{wrapfig}

\title{\LARGE \bf
Music-Driven Legged Robots: Synchronized Walking to Rhythmic Beats
}

\author{Taixian Hou$^{1}$, Yueqi Zhang$^{1}$, Xiaoyi Wei$^{1}$, Zhiyan Dong$^{1}$, Jiafu Yi$^{2}$, Peng Zhai$^{1,*}$, Lihua Zhang$^{1,*}$
\thanks{$^{1}$Taixian Hou, Yueqi Zhang, Xiaoyi Wei, Zhiyan Dong, Peng Zhai, Lihua Zhang are with the Academy for Engineering and Technology, Fudan University, Shanghai 200433, China.
         {\tt\small txhou21@m.fudan.edu.cn;}}%
\thanks{$^{2}$Jiafu Yi is with the School of Information and Communication Engineering, Hainan University, Foshan 200433, China.}
\thanks{*Corresponding Author.}
\thanks{The work reported in this paper was supported by the National Key R\&D Program of China (Grant Number: 2021ZD0113502, 2021ZD0113503),  China Postdoctoral Science Foundation (Grant Number: BX20220071, 2022M720769), and Research on Basic and Key Technologies of Intelligent Robots (Grant Number: KEH2310017).}
}

\let\oldtwocolumn\twocolumn
\renewcommand\twocolumn[1][]{%
    \oldtwocolumn[{#1}{
    \begin{center}
           \vspace{-0.3cm}
           \includegraphics[width=\textwidth]{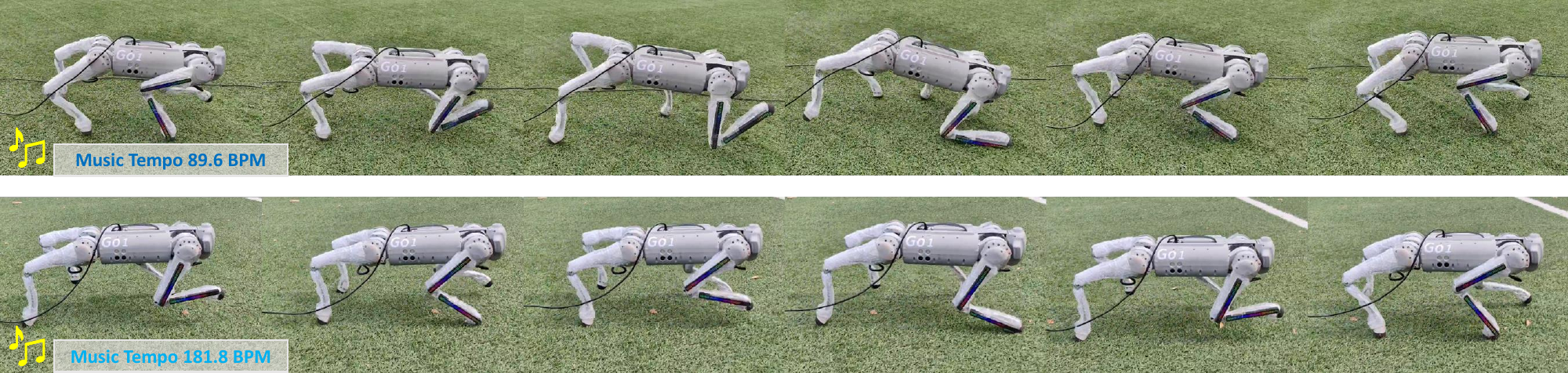}
           \vspace{-0.5cm}
           \captionsetup{font=small}
           \captionof{figure}{Sim2Real Implementation of the Music-Driven Controller on the Go1 Robot. We tested the controller's real-time performance and stability in outdoor environments. The quadruped robot achieved rhythm-synchronized motions in response to mLO1 (89.6 BPM) and mJB1 (181.8 BPM) music tracks from the AIST++ dataset, demonstrating two distinct gaits aligned with the different rhythmic tempos. The right front (RF) leg, highlighted with light strips, moved in sync with the music. Video is available at https://music-walker.github.io/.}
           \label{fig::deployment1}
        \end{center}
    }]
}

\begin{document}
\maketitle

\thispagestyle{empty}
\pagestyle{empty}

\begin{abstract}
We address the challenge of effectively controlling the locomotion of legged robots by incorporating precise frequency and phase characteristics, which is often ignored in locomotion policies that do not account for the periodic nature of walking. We propose a hierarchical architecture that integrates a low-level phase tracker, oscillators, and a high-level phase modulator. This controller allows quadruped robots to walk in a natural manner that is synchronized with external musical rhythms. Our method generates diverse gaits across different frequencies and achieves real-time synchronization with music in the physical world. This research establishes a foundational framework for enabling real-time execution of accurate rhythmic motions in legged robots. Video is available at website: https://music-walker.github.io/.
\end{abstract}

\section{INTRODUCTION}

Animal locomotion behaviors often exhibit strong periodicity, such as leg movements of mammals during walking and the wing flapping of birds. These behaviors are closely linked to the nervous systems of animals\cite{RN73}. The periodicity of these behaviors overlaps with the rhythmic patterns found in music. In fact, rhythm synchronization is a rare but notable phenomenon observed in both humans and animals. Humans exhibit this behavior from early infancy\cite{RN59}, where studies have shown that even babies move rhythmically to musical beats\cite{RN83}. 

Among animals, birds, particularly parrots such as the well-known cockatoo "Snowball," have demonstrated the ability to synchronize movements with musical beats and even adjust to varying tempos\cite{RN84}. Bonobos have been observed drumming to a musical beat\cite{RN85}, and chimpanzees can follow simple rhythms\cite{RN86}. Some marine mammals, such as California sea lions, have demonstrated head-bobbing in sync with musical beats, though these behaviors are rare and usually the result of extensive training\cite{RN87}. This suggests that the ability to synchronize movements to musical rhythms is influenced by learned behaviors\cite{RN62}.

Reinforcement learning parallels how animals develop skills through training and has proven to be effective in control of legged robots.\cite{RN23, RN16}. However, most learning controllers for legged robots respond primarily to velocity commands, often neglecting the importance of gait frequency and phase. Margolis, et al.\cite{RN49} proposed a learning controller capable of following multiple command signals, including specific gait frequencies.

In model-based approaches, Central Pattern Generator (CPG) models are frequently employed for controlling legged robots\cite{RN89, RN88}. CPG models provide stable periodic trajectories, enabling smooth locomotion. Recently, hybrid approaches that combine CPG models with reinforcement learning have emerged\cite{RN58, RN74}. A recent approach designs a policy that outputs several parameters to control the foot trajectory. Joint positions are then computed using inverse kinematics, then tracks through PD control\cite{RN63, RN66}.

Our approach differs from these methods in several key aspects, particularly in how we combine a low-level phase tracker with an oscillator for more precise gait control. This will be discussed in detail in the following sections. The main contributions of this paper are as follows:

1)We propose a combination of a trainable phase tracker and oscillator that can accept nominal frequency commands, allowing for precise frequency tracking in legged robots.
2)We introduce a ground reaction force prediction module that allows deployment on robotic platforms without the need for high-precision force sensors. The initial setup enables the phase tracker to quickly converge to stable gaits.
3)We present a hierarchical control framework suitable for real-time operation on legged robots, enabling seamless transfer from simulation to reality. This controller dynamically adjusts gait patterns to synchronize with external musical rhythms while maintaining forward locomotion. Our work not only adds entertainment value to robotics but also provides a level of bio-inspired functionality.

\section{PRELIMINARIES and RELATED WORK}

We Introduce music data processing and phase oscillators.

\subsection{Music Data Processing}

Rhythmic tempo in music typically measures the periodicity of high-intensity signals, with frequencies closely aligned to animal behavior. For instance, 120 to 140 beats per minute (BPM) is commonly found in electronic dance music and can be easily perceived and synchronized with. In most cases, the tempo remains constant, although some music may exhibit minor tempo variations.

\begin{figure}[h]
  \centering
  \includegraphics[width=0.47\textwidth]{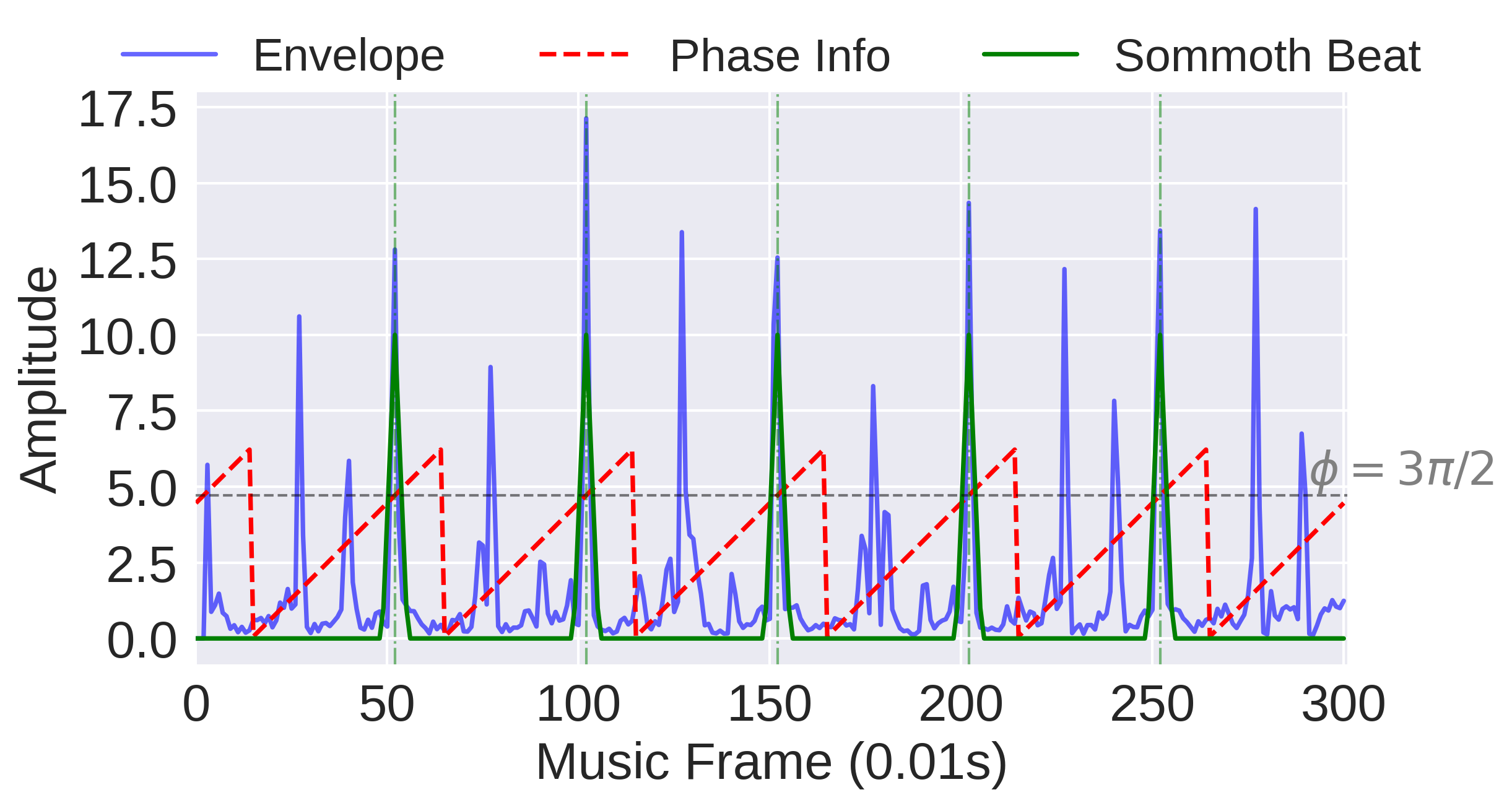}
  \captionsetup{font=small}
  \caption{Visualization of music features sampled at 100 Hz, including the envelope (music intensity), smoothed beats, and interpolated phase information, based on a 3-second music clip with a tempo of 120 BPM.}
  \label{fig:2}
  \vspace{-0.1cm}
\end{figure}

Fig. \ref{fig:2} provides a visualization of the music features. The green vertical dash-dotted lines indicate the beats in the music, which have been smoothed using a one-dimensional convolution to generate the green curve. This smoothed beat data is utilized in Eq. (\ref{eq:9}) "Reward1" for comparisons.

As described in the following sections, the robot's dynamic footfall phase is set to occur at $\frac{3\pi}{2}$. To achieve alignment between the robot’s footfall and the music beats, we match the robot's foot contact with the music beat at this specific phase value of $\frac{3\pi}{2}$. We perform linear interpolation within each beat cycle to assign phase information to each music frame, which is then utilized in our reward functions.

\subsection{Robot Control with Phase Oscillators}

The oscillator forms the core of phase generation in quadruped robots. Our oscillator systems can be understood as special cases of the following general system:

\begin{equation}
\begin{aligned}
\dot{\phi}_{i} &= \tilde{\omega} + \delta(\phi_{i}, x_{i}) \\
u &= f(\phi_{i}, x_{i})
\end{aligned}
\end{equation}

where $i \subseteq \{1, \ldots, n\}$, represents the quadruped robot oscillator index for each leg ($n=4$). Here, $\phi$ represents the oscillator phase, $\tilde{\omega}$ is the intrinsic frequency, and $\delta()$ is a frequency modulator that typically follows the dynamics of the oscillator.  $f()$ acts as the control function, which is determined by specific phase $\phi_i$ and observations $x_i$ of the robot, and output actions $u$.

When $\delta() = 0$, the system simplifies to a fixed-frequency clock\cite{RN72}. This setup, as used by Shafiee, et al.\cite{RN63} and Siekmann, et al.\cite{RN61}, places the burden of oscillator dynamics on the design of $f()$.

Classical CPG systems are divided into those without feedback and those with feedback. In the case of no feedback, $\delta() = \delta(\phi_i)$, meaning the robot’s motion adheres to a pre-defined dynamic pattern. This setup provides stable, continuous, self-excited oscillations with properties like limit cycle convergence, as seen in Kuramoto\cite{RN90} and Hopf oscillators\cite{RN65}. When observations are added, i.e., $\delta() = \delta(\phi_i, x_i)$, the system not only captures dynamic characteristics but also adjusts the dynamics based on the robot’s current state. This introduces a highly adaptive mechanism.

Our method incorporates this setup, particularly the feedback model proposed by Owaki, Kano, et al.\cite{RN60}. This setup can also be interpreted as a feedback-decentralized oscillator system, where ground reaction forces (GRF) are used to synchronize the oscillators. Such structures, as employed by Ryu and Kuo\cite{RN91}, exploit the advantages of various oscillators while granting each leg a degree of autonomy.

The differential equation for oscillator is:

\begin{equation}
\dot{\phi}_i = \omega - \sigma N_i \cos \phi_i,
\end{equation}

Due to the instability of the ground reaction force measurements, this approach is difficult to apply. Zhang, Heim, et al.\cite{RN68} proposed replacing the original force $N_i$ with the normalized ground reaction force $G_i^{\text{norm}}$. This replacement not only makes it easier for the policy to learn, but also helps to overcome the instability of force sensors in the IsaacGym simulator. Additionally, by introducing a bias term $\xi$, the stability problem in maintaining a stationary posture without velocity commands is resolved.

However, in this approach, the oscillator's frequency is directly bound to the velocity command, which prevents the quadruped robot from accepting external nominal frequency inputs. To address this issue, during the low-level network training, we applied domain randomization to generate more diverse combinations of velocity and frequency. Whether the low-level network can capture this complexity is crucial for achieving beat synchronization, which will be examined in the experimental section. Another challenge arises after the robot has stabilized in a stationary stance: due to the lack of an initial phase intervention, the relative phase difference (RPD)  between oscillators relies on ground reaction feedback $G_i^{\text{norm}}$ to "guide" the robot into stability. This results in a delay in the gait convergence to the optimal pattern, such as the trot gait. The robot may start with a bound gait and take several seconds to converge or even exhibit erratic behavior in an attempt to adjust the phase difference more quickly.

\begin{figure}[h]
  \centering
  \includegraphics[width=0.47\textwidth]{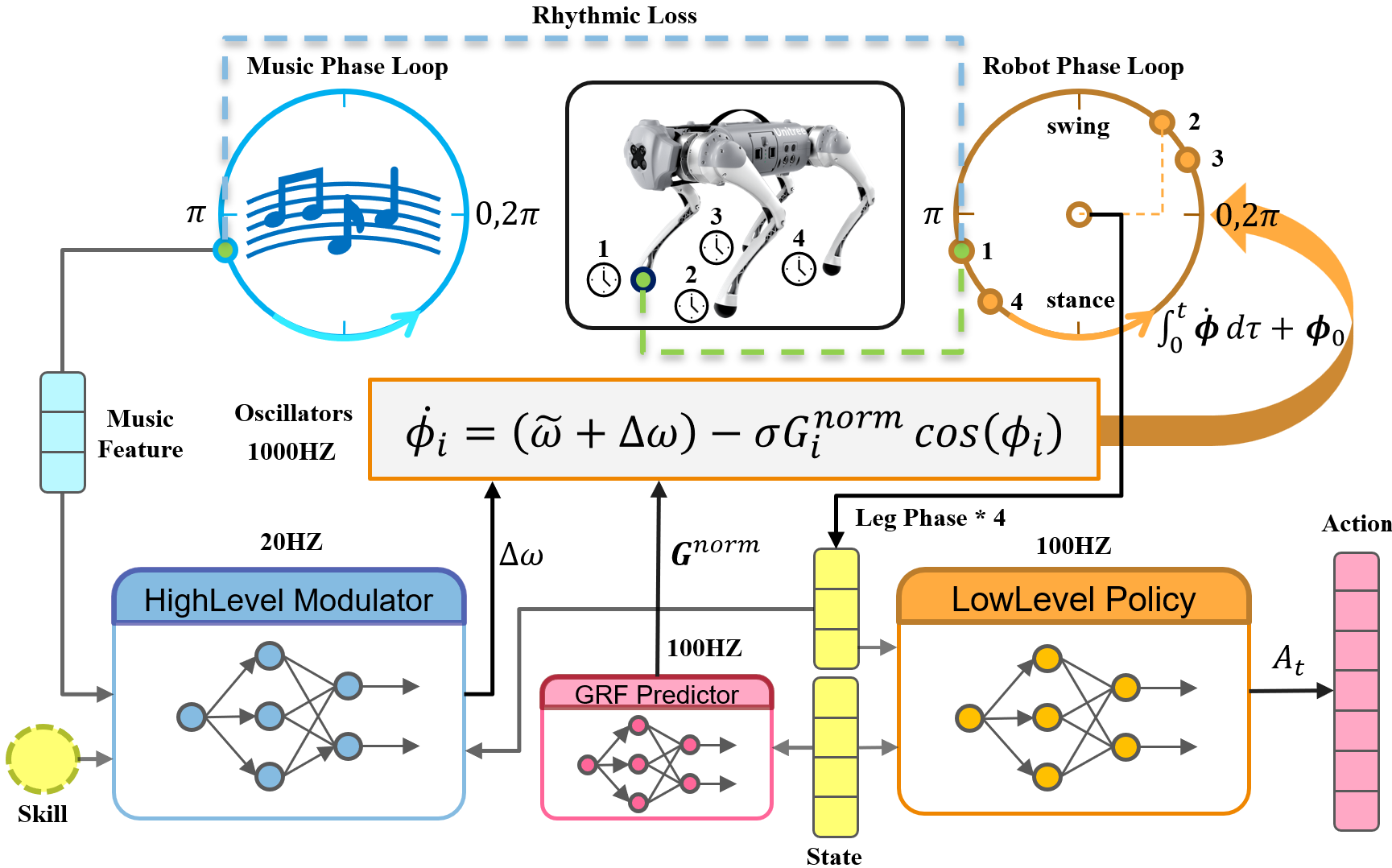}
  \captionsetup{font=small}
  \caption{Proposed controller, features a hierarchical structure with three main modules: the high-level Phase Synchronization Modulator, the low-level Phase Tracking Policy, and the Distributed Phase Oscillators. In the figure, the top-left and top-right corners respectively depict the music phase ring and the robot's oscillator phase ring, both with a range of $0$ to $2\pi$. The green dot on the music phase ring represents the current music frame, and its features (such as beat and phase observations) are sent to the high-level network. The four points on the robot's phase ring correspond to the phases of the oscillators bound to each of the four legs. Oscillator Num.1 (associated with the right front leg) is tasked with synchronizing with the musical beat.}
  \vspace{-0.1cm}
  \label{fig:3}

\end{figure}

\section{METHOD}

\subsection{Overview} 

Inspired by animals in nature, we designed a controller that enables a quadruped robot to follow musical rhythms. Specifically, the robot is programmed such that one selected leg experiences the maximum ground contact force during the gait cycle precisely when a musical beat occurs. Achieving this synchronization requires meeting two consistency conditions: \textbf{frequency consistency} and \textbf{phase consistency}. Additionally, we aim to ensure that the robot's movements remain smooth and natural, without introducing abrupt or unnatural jerks to lock onto the beat. 

To address the above requirements, we propose a hierarchical controller, as illustrated in Figure 3. In this architecture, each level of the network targets one of the specific consistency conditions. This structured approach enables our controller to exhibit behaviors natural movements while achieving precise musical synchronization. The training is divided into two distinct stages.

\subsection{Stage 1: Training Low-level Phase Tracker Network with Oscillators to Meet Frequency Consistency:}

In the first stage, we focus on training the low-level network, where frequency modulation is achieved by cascading oscillators with the trained low-level policy. The network can be viewed as a phase tracker, denoted by $\pi_l(a|s,\phi)$, where $a$ denotes actions specified by joint target positions, $s$ consists of the velocity command $v_{cmd}$, the robot’s proprioceptive state information $o$, and the previous action $a_{last}$. 

\subsubsection{\textbf{Network Observations}}

Angular velocity of the base, gravity vector, velocity commands, positions and velocities from 12 joints, phase observations of the oscillators $\phi_{obs} \in \mathbb{R}^8$, and the actions from the previous step, forming a 53-dimensional observation space.

\subsubsection{\textbf{Special Reward}}
The internal phase-based reward ties the phase with the leg's swing and stance actions. Combined with the oscillator’s dynamic properties, this reward enables the robot to achieve continuous internal phase awareness and learn the nominal internal frequency $\tilde{\omega}$. As shown in Fig. \ref{fig:loop}(a), this reward penalizes ground reaction forces when $\phi_i \in [0, \pi)$, marking this phase as the swing state. Conversely, it rewards ground reaction forces when $\phi_i \in [\pi, 2\pi)$, designating this phase as the stance state. It is important to note that this reward is conceptually simpler and clearer, ensuring reliable learning. Along with the oscillators, it commits gait continuity.

\begin{equation}
    r_{\mathrm{Phase}} = - \sum_{i=1}^{4} G_i^{\mathrm{norm}} \sin{(\phi_i)}
\end{equation}

\subsubsection{\textbf{Oscillator Design}}

For the quadruped robot, we configure four oscillators corresponding to the legs in the order of right front (RF), left front (LF), right hind (RH), and left hind (LH) leg. 
The phase observation for each leg can be expressed as $\phi_i(t) = \int_0^t \dot{\phi}_i(t) dt + \phi_i(0)$, where $\dot{\phi}_i(t)$ is the phase rate of change, and $\phi_i(0)$ is the initial phase at startup. The rate of phase change is governed by the function $\dot{\phi}_i(t) = f(\tilde\omega, G_i^\mathrm{norm}, \phi_i(t))$, which incorporates a feedback model that accounts for the ground reaction forces. 

Ultimately, our oscillator can be expressed as: 

\begin{equation}
\phi_i(t) = \int_0^t (\tilde{\omega} - \sigma G_i^{\text{norm}} (\cos \phi_i(t) + \xi)) \, dt + \phi_i(0)
\end{equation}

\begin{figure}[!h]
  \centering
  \includegraphics[width=0.45\textwidth]{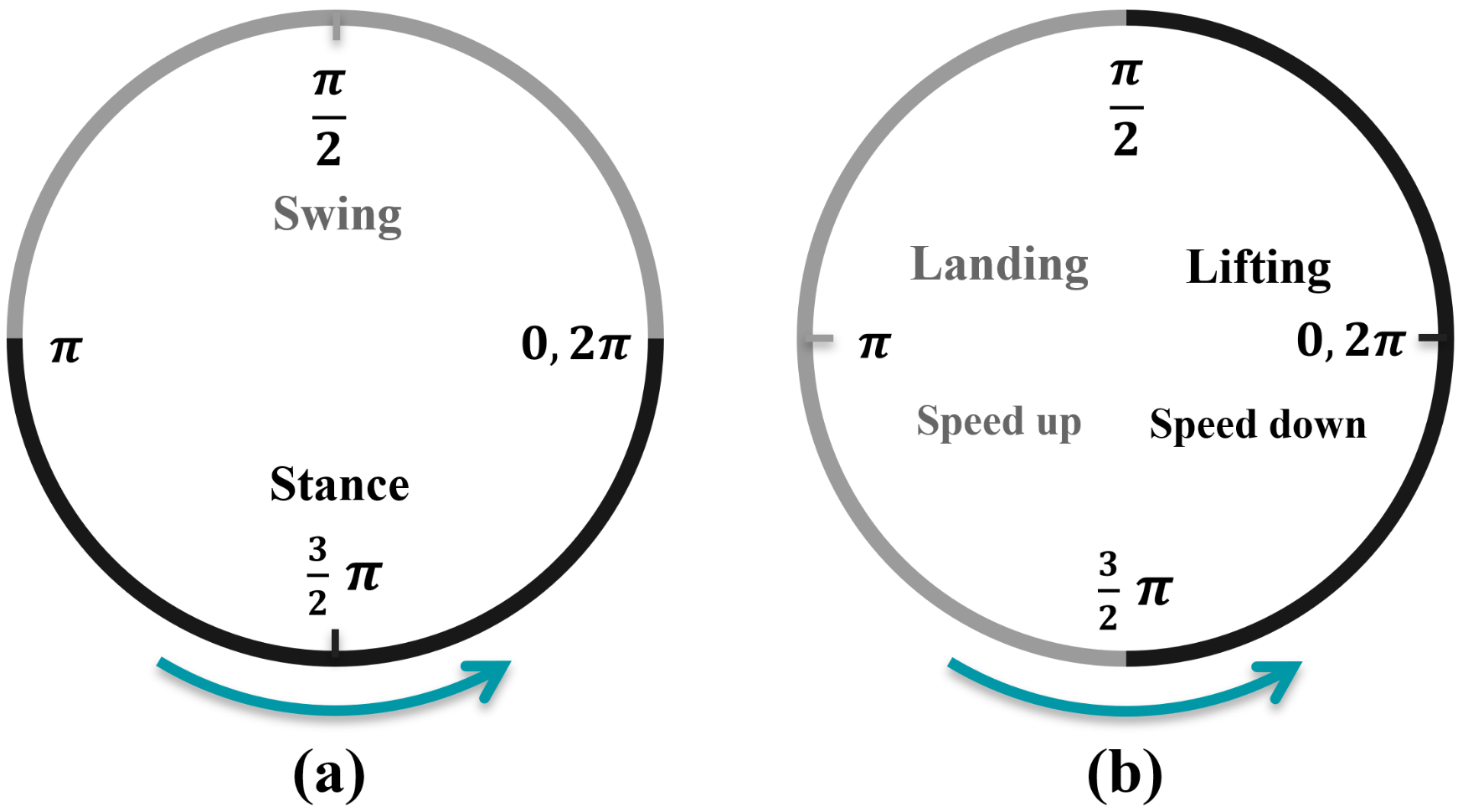}
  \captionsetup{font=small}
  \caption{Robot’s oscillator phase loop within range [0, 2$\pi$].}
  \label{fig:loop}
  \vspace{-0.2cm}
\end{figure}

As shown in Fig. \ref{fig:loop}(b), influenced by the term $\cos \phi_i(t)$, $\phi_i \in [\frac{\pi}{2}, \frac{3\pi}{2})$ represents a acceleration of the phase change at landing, while the right half of the phase ring represents an deceleration at lifting. 

\begin{equation}
[\tilde{\omega}, \sigma, \xi, \phi_i(0)] =
\begin{cases} 
[1, 4, 1, \frac{3\pi}{2}] & \text{if } |v_x| \leq 0.5 \\
[\omega, 2\pi, 0, \frac{\pi}{2}] & \text{if } i \in M \\
[\omega, 2\pi, 0, \frac{3\pi}{2}] & \text{otherwise}
\end{cases}
\label{eq:5}
\end{equation}

where $G_i^{\text{norm}} = \min\left(\frac{N_i}{g \cdot \text{Mass}}, 1.0\right)$, $\tilde{\omega}$ is the intrinsic angular frequency, $\omega = 2\pi f$ is the commanded angular frequency, with $f \in (1.0, 4.0)$, and $\xi$ is the offset. $M$ represents the set of symmetric legs with lower load\cite{RN71}.

\begin{equation}
M = 
\begin{cases} 
[1,4] & \text{if } (N_1 + N_4) < (N_2 + N_3)\\ 
[2,3] & \text{otherwise} 
\end{cases}
\end{equation}

\subsection{Ground Reaction Force Estimator}

In practical scenarios, the normalized ground reaction force $G^{\text{norm}}$ in the oscillator formula is often constrained by hardware limitations. One approach is to set it as a constant $a=0.25$, which degrades the oscillator to a feedback-free form. Alternatively, we can predict the normalized ground reaction force based on proprioceptive observations that include the foot contact indicator $I$ and joint information $s$. This allows us to maintain the advantages of a feedback oscillator. We use the simulated ground reaction forces provided by IsaacGym as ground truth labels to train this network, which operates at a frequency of 100 Hz.

To enable the policy to better adapt to the predicted reaction forces, we employ an online curriculum-like learning approach with loss $\text{MSE}(G^{\text{sim}}, G^{\text{pred}}(s, I))$, where a mixture of simulated and predicted reaction forces replaces the oscillator's reaction force formula. The mixed reaction force is controlled by a weight coefficient $\rho = \frac{\text{iteration}}{N}$.

\begin{equation}
G^{\text{norm}} = \min \left((1 - \rho) \times G^{\text{sim}} + \rho \times G^{\text{pred}}(s, I), 1.0 \right)
\end{equation}

Where $N$ is the total number of training iterations. This gradual approach avoids directly using untrained predicted feedback forces, which could lead to oscillator instability.

\subsection{Stage 2: Training High-level Phase Modulator to Meet Phase Consistency:}

After the first stage of training, we developed a phase-tracking policy capable of following nominal frequency commands within a specific range. To enable the robot to smoothly adjust its internal phase and achieve beat synchronization, we retained the low-level network and froze its parameters while training the high-level network. The high-level phase modulator $\pi_h(\delta\omega|\omega_m, \theta_m, \phi)$ receives musical features, including the beat frequency $\omega_m$ and the current phase of the musical frame $\theta_m$, derived from a short-time analysis of the music. It adjusts the robot’s internal frequency $\tilde\omega = \omega_m + \delta\omega$ through the intervention of $\delta\omega$ to correct the phase, simplifying to $\pi_h(\tilde\omega|\omega_m, \theta_m, \phi)$.

Overall, the robot’s gait frequency is approximately synchronized with the music. Our hierarchical controller $\pi(a|s, \omega_m, \theta_m):=\pi_l(a|s, \phi)\phi(\tilde\omega, t)\pi_h(\tilde\omega|\omega_m, \theta_m, \phi)$, generates walking motions that align with both the musical features and the robot’s proprioceptive state. This policy integrates both the frequency and phase of the music, fulfilling the necessary conditions for achieving beat synchronization. As illustrated in Fig. \ref{fig:3}, the phase spaces of both the music phase loop and the robot's oscillator phase loop are within the $[0, 2\pi]$ range. To ensure continuity in phase observation, each phase observation is represented in 2D as $\phi_{obs} = (\cos\phi, \sin\phi) \in \mathbb{R}^2$. The goal of the phase modulation network is to minimize the Rhythmic Loss between the phase points on the two loops.

The structure of the high-level network is as follows:

1) \textbf{Observations:} Rhythm tempo extracted from the music window, phase observation of current music frame, and oscillators' phase observations, totaling 11 dimensions.

2) \textbf{Output:} The network output provides the internal frequency adjustment value $\delta \tilde{\omega}$ which was queried at a frequency of 20 Hz. This value was then used to calculate the internal frequency $\tilde\omega = \omega_m + \delta\omega$. 

3) \textbf{Special Reward:} The rhythm consistency reward is essential for the successful learning of the high-level network. Since the low-level network has already been trained to track phases, we fully leverage this capability by directly transforming rhythm consistency into the dynamic elimination of phase differences. The reward is described as follow:

\begin{equation}
r_{\text{Rhythm}} = \exp\left(-\sigma(\phi_{obs,j}^2 - \theta_{obs}^2)\right)
\end{equation}

where $\sigma$ is the scaling coefficient, and $j$ represents the oscillator that needs to be synchronized with the music. In our experimental setup, the music is bound to oscillator Num.1 corresponding to the \textbf{RF leg}. The squared difference between $\phi_{obs}$ and $\theta_{obs}$ can be understood as the squared geometric distance between the phases on the unit rings.

\subsection{Training configuration Details}

All training in this work was accelerated using the Isaac Gym simulator\cite{RN70}, with parallel training conducted across 4096 environments on an NVIDIA RTX 4090. During the training of the high-level network, we utilized the AIST++ dataset. Initially, all music tracks in the dataset were concatenated into a continuous sequence. Subsequently, 50,000 random 10-second music segments were pre-loaded, from which the robot sampled during training in each environment.

\section{EXPERIMENTAL RESULTS and DISCUSSION}

\begin{figure*}[t]
    \centering
    \includegraphics[width=\linewidth]{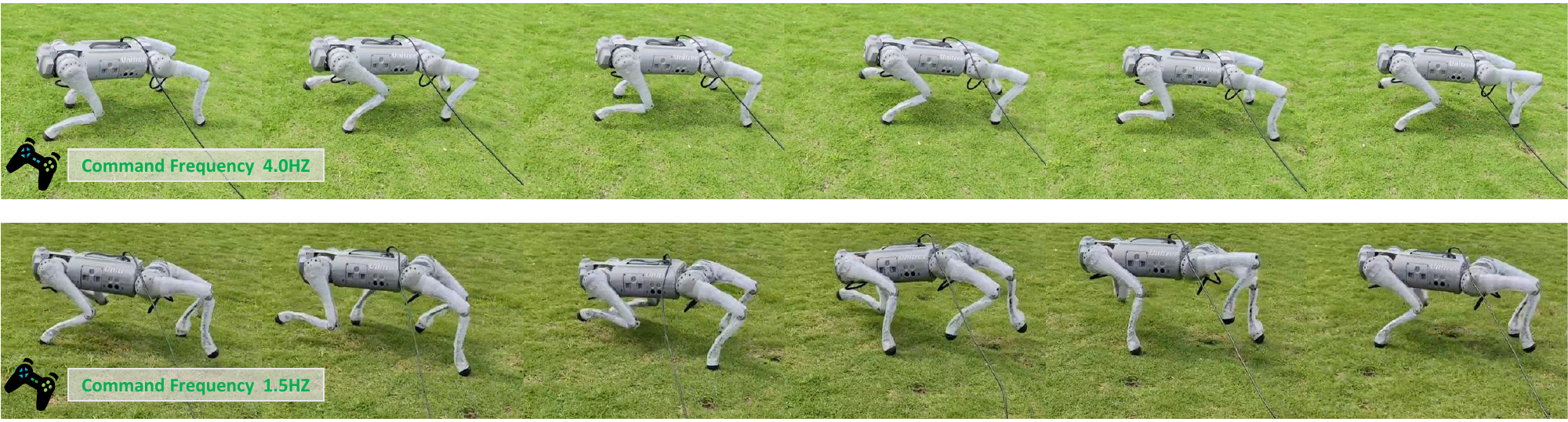}
    \captionsetup{font=small}
    \caption{Frequency tracking Sim2Real Experiments conducted on Go1 robot with Low-Level Networks and Oscillators. Our controller demonstrates diverse gait skills. The robot experiment videos are recorded at 60 fps. Here, we have extracted five images from videos of 4.0 Hz and 1.5 Hz frequencies commands, with each image sampled every 10 frames.}
    \label{fig:fwalk}
    \vspace{-0.2cm}
\end{figure*}

Our method was applied to the Unitree Go1 robot, conducting both simulation and hardware experiments. 

\subsection{Frequency Command Tracking Experiment of the Low-Level Network}

In this experiment, we only utilized the low-level phase tracker policy and the oscillator modules. We tested performance on flat terrain in the IsaacGym simulator. To obtain more precise frequency data, we increased the sampling rate of force and base information from 100 Hz during training to 500 Hz in these tests.

\begin{figure}[t]
    \centering
    \begin{minipage}{0.23\textwidth}
        \centering
        \includegraphics[width=\linewidth]{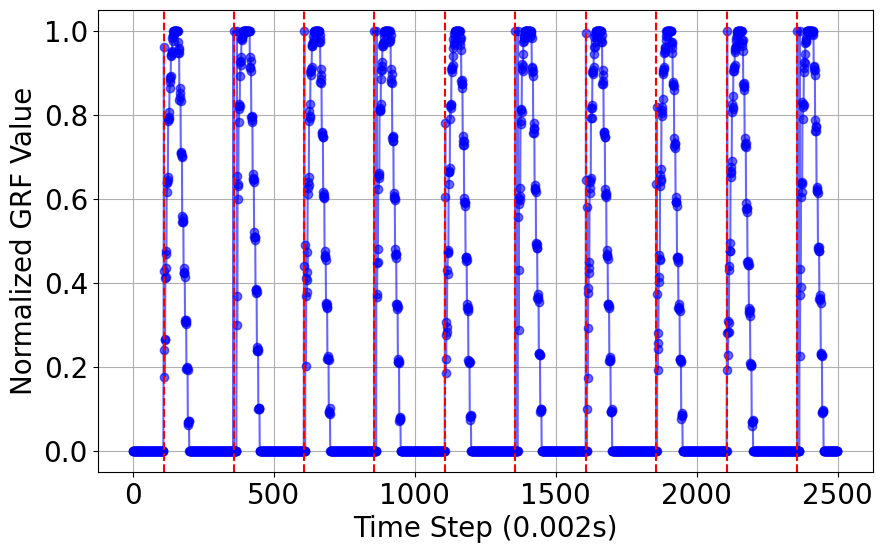}
        (a)
    \end{minipage}
    \hfill
    \begin{minipage}{0.23\textwidth}
        \centering
        \includegraphics[width=\linewidth]{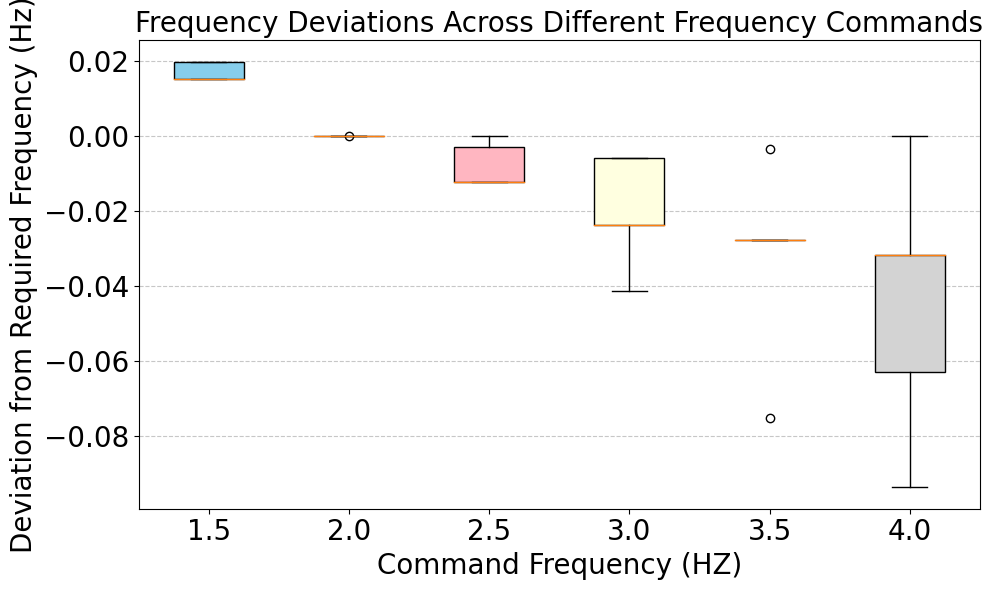}
        (b)
    \end{minipage}
    \hfill
    \captionsetup{font=small}
    \caption{Ground reaction force (GRF) data collected over 2,500 time steps (5 seconds). (a) Shows normalized GRF values at a command frequency of 2.0 Hz, highlighting clear periodic patterns. (b) Displays frequency deviation data across six different command frequencies.}
    \label{fig:6}
    \vspace{-0.3cm}
\end{figure}

The robot was commanded to move forward at a velocity of $V_{cmd} = 0.8 \text{m/s}$ with different nominal frequency combinations. We tested six sets of nominal frequency commands $f_{cmd}$ within the range [1.5, 4.0] Hz, with a 0.5 Hz increment between each. 
We analyzed the actual stepping frequency from the ground reaction force (GRF) data. As shown in Fig \ref{fig:6}(a), we present a segment of trajectory data from the right front leg (RF) of the robot. The GRF exhibits a periodic pattern, with the red dashed vertical lines indicating the first upward moment of each cycle, corresponding to the robot’s foot contact with the ground. By measuring the time interval between consecutive foot contacts, we can calculate the single-step cycle for the leg. 

\begin{figure}[t]
    \centering
    \begin{minipage}{0.23\textwidth}
        \centering
        \includegraphics[width=\linewidth]{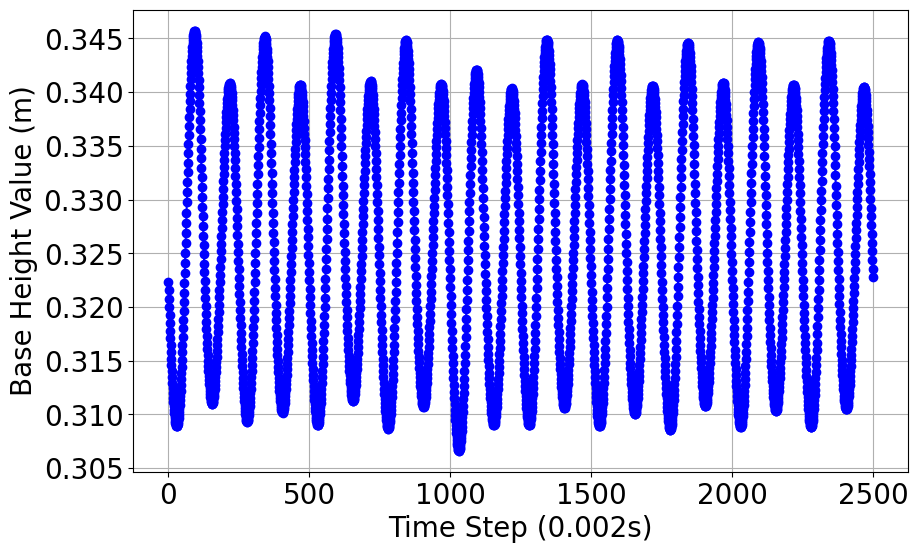}
        (a)
    \end{minipage}
    \hfill
    \begin{minipage}{0.23\textwidth}
        \centering
        \includegraphics[width=\linewidth]{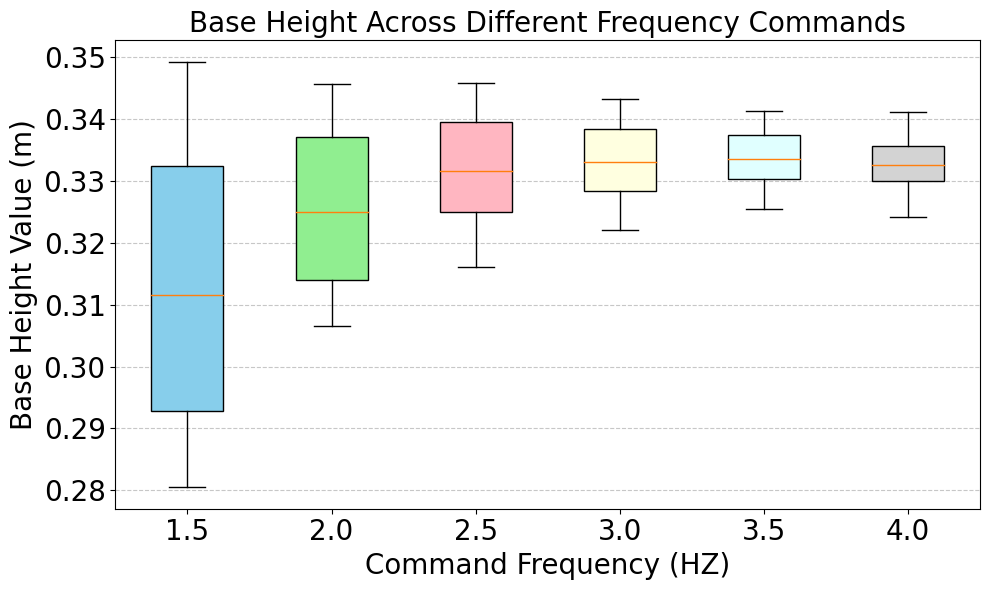}
        (b)
    \end{minipage}
    \hfill
    \captionsetup{font=small}
    \caption{Base height data collected over 2,500 time steps (5 seconds). (a) Displays base height values at a 2.0 Hz command frequency, demonstrating good base stability. (b) Shows the range of base height fluctuations decreases with increasing gait frequency, highlighting the impact of frequency on stability.}
    \label{fig:7}
    \vspace{-0.3cm}
\end{figure}

We computed the difference between the actual stepping frequency and the commanded frequency under different frequency commands. The results in Fig. \ref{fig:6}(b) indicate average frequency deviations less than 0.05 and variance under 0.01 for all tested frequencies. This experiment demonstrates the strong performance of the low-level network, thus providing a stable frequency-tracking foundation for the high-level network to achieve rhythm synchronization. We also collected base height data in Fig. \ref{fig:7} to study the impact of the robot's gait frequency on stability, showing that base stability increases with higher frequencies.

\subsection{Rhythm Synchronization Experiment}

Since the walking process involves interaction with the ground, we can consider the moments of peak ground reaction force (GRF) within a gait cycle as the robot’s kinematic beats. In research on dance generation, the synchronization of virtual avatars' dance motions with the music beat is frequently studied\cite{RN78, RN79, RN76, RN77}. Our music data comes from the AIST++ dataset\cite{RN75}.

In this experiment, we introduce two additional reward functions to promote beat synchronization for comparison. The first reward, applies a one-dimensional convolution to the beat, producing a temporal smoothed beat $\mathcal{B}(t)$, represented by the green line in Fig \ref{fig:2}. This formulation applies a reward based on the phase difference between the oscillator’s internal phase $\phi_{\text{osc}}(t)$ and the reference stance phase when a beat occurs.

\begin{equation} 
r_1(t) = \mathcal{B}(t) \cdot \exp\left( - \left( \phi_{\text{osc}}(t) - \frac{3\pi}{2} \right)^2 \right)
\label{eq:9}
\end{equation}

The second reward, inspired by \cite{RN78}. $r_2(t)$ penalizes the absence of gait beats during a musical beat:

\begin{equation}
r_2(t) = 
\begin{cases}
-1, & \text{if } \exists \text{music beat} \land \nexists \text{kinematic beats} \\
1, & \text{otherwise}
\end{cases}
\end{equation}

All experiments were conducted using the same velocity command ($V_{cmd} = 0.8m/s$). We selected three music tracks from the AIST++ dataset and tested three high-level networks each trained with different reward functions. The following two outcomes were of primary interest:

Beat Alignment: Calculated time difference $\Delta t$ between kinematic beats and musical beats, indicating the synchronization performance between robot gait and music.

Intrinsic Frequency Variance: We aimed for minimal adjustments to the internal frequency to achieve phase alignment. Excessive frequency changes can result in unnatural jerks in the robot’s gait. Therefore, we monitor the intrinsic frequency data.

\newcommand{\mytab}{\centering\arraybackslash} 
\newcommand{\mean}[1]{\multicolumn{1}{c}{#1}} 

\begin{table}[htbp]
\vspace{0.2cm}
\centering
\begin{tabular}{|c|c|c|c|c|c|c|}
\hline
\multirow{2}{*}{\textbf{Music (BPM)}} & \multicolumn{2}{c|}{\textbf{Ours}} & \multicolumn{2}{c|}{\textbf{Reward1}} & \multicolumn{2}{c|}{\textbf{Reward2}\cite{RN78}} \\ \cline{2-7}
& \textbf{$\Delta t_{max}$} & \textbf{$\sigma(\tilde \omega)$} & \textbf{$\Delta t_{max}$} & \textbf{$\sigma(\tilde \omega)$} & \textbf{$\Delta t_{max}$} & \textbf{$\sigma(\tilde \omega)$} \\ 
\hline
mLO1 (89.6) & \textbf{0.06} & \textbf{0.045} & 0.45 & 1.925 & 0.23 & 0.707 \\ \cline{2-7}
mJS4 (120.0) & \textbf{0.01} & \textbf{0.040} & 0.04 & 3.832 & 0.08 & 0.782 \\ \cline{2-7}
mJB1 (181.8) & \textbf{0.03} & \textbf{0.043} & 0.05 & 2.911 & 0.06 & 1.172 \\ \cline{2-7}
\hline
\end{tabular}
\captionsetup{font=small}
\caption{Performance metrics for different reward functions across various music BPM settings. The table compares ours with Reward1 and Reward2. The lowest values are highlighted in bold.}
\label{performance}

\end{table}

The results are presented in Table \ref{performance}.  All three rewards achieved strong beat synchronization with low time difference. However, the two additional rewards led to greater frequency adjustments. In contrast, our controller, which rewards directly based on the "distance" between two phase rings, yielded the best beat alignment with significantly lower frequency variance.

\begin{figure}[t]
    \centering
    \begin{minipage}{0.23\textwidth}
        \centering
        \includegraphics[width=\linewidth]{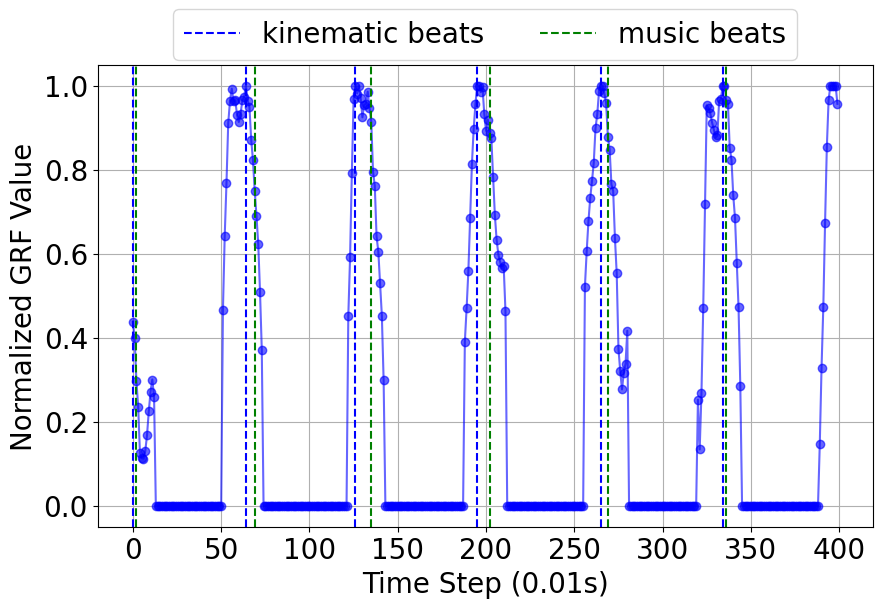}
        (a)
    \end{minipage}
    \hfill
    \begin{minipage}{0.23\textwidth}
        \centering
        \includegraphics[width=\linewidth]{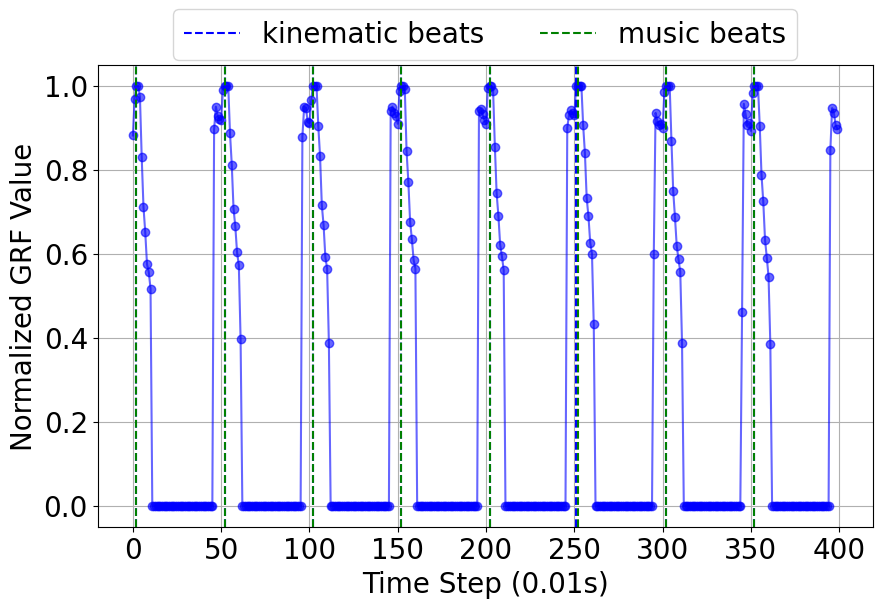}
        (b)
    \end{minipage}
    \hfill
    \captionsetup{font=small}
    \caption{Beats Alignment between Music and Robot Gait. Here we visualize the kinematic beats (blue dotted line) of our robot gait, and music beats (green dotted line). The kinematic beats are extracted by finding local maxima in GRF periods.}
    \label{fig:syn}
    \vspace{-0.3cm}
\end{figure}

Fig. \ref{fig:syn} demonstrates data from our controller under the "mLO1" and "MJS4" music track. We present a 4-second segment of trajectory data. The blue line indicates the maximum ground reaction force during each gait cycle, which represents the kinematic beat. The green line marks the musical beat. As shown, the controller achieves a strong correlation between the robot's action and the music.

\subsection{Sim2Real Experiment}

\subsubsection{Real-time Challenge}

A major challenge for real-world deployment is dealing with latency. While in simulation, the controller operates in an ideal environment without noise or delays. In reality, several factors contribute to latency, including: inference time of the neural network policy, the transmission latency of ROS2 topics, the delay in the microphone acquisition, and the time consumed in feature extraction within the music window are all sufficient to cause failure in this task, which requires strict temporal synchronization. To mitigate these issues, we implemented the following solutions:

\textbf{Balancing the Music Frame Window Size:} The window must be large enough to extract stable beats while being small enough to enable faster execution.

\textbf{Accurate phase calculation:} We timestamp each signal collected by the microphone and compute the music phase based on the current time before the high-level network inference. This approach ensures accurate music phase estimation even with intermediate delays.

\textbf{Frequency smoothing:} Sudden noise spikes can disrupt music analysis. We implemented a frequency queue of size 5 and applied a median filter to smooth frequencies.

\subsubsection{Settings}

\begin{wrapfigure}{r}{0.23\textwidth}
  \centering
  \includegraphics[width=0.22\textwidth]{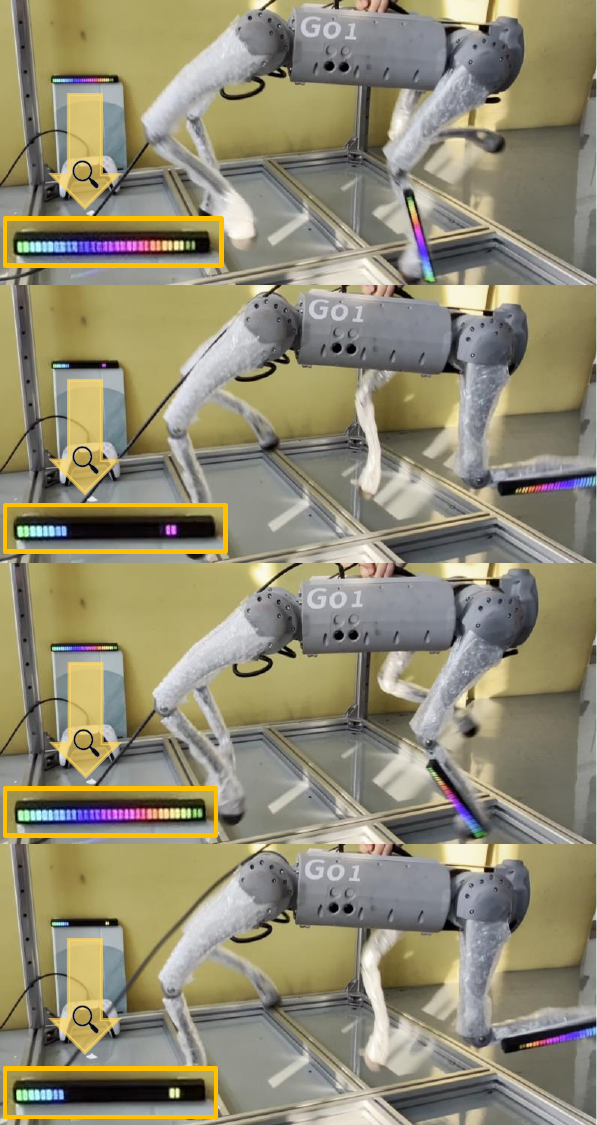}
  \captionsetup{font=small}
  \caption{Key frames from a suspension synchronization experiment. The light strip on the left is placed near the sound source to visualize the music's beat.}
  \label{fig:hang}
  \vspace{-0.2cm}
\end{wrapfigure}

Our hierarchical controller was deployed on the Unitree Go1 robot. The robot control software runs on a host machine using a ROS2 C++ interface and interacts with the robot via an Ethernet connection. The trained joint policy is exported as TorchScript. The low-level phase tracking and ground reaction prediction networks are queried at a frequency of 100Hz, while the high-level frequency modulator runs at 20Hz, and the phase oscillators are set to a frequency of 1000Hz. A PS5 controller serves as the microphone input device.

\subsubsection{Results}

a) Suspension Experiment: As shown in Fig. \ref{fig:hang}, to capture both the robot's motion beats and the musical beats from a fixed camera position, we suspended the robot. The light strip on the left is used to detect the music beats, while the one on the right is bound to the \textbf{RF} leg to emphasize its movement, which is required to be synchronized with the music. This setup enabled us to visualize the synchronization between the kinematic beat (footfall) and the music beat.

b) Outdoor Experiment: In comparison to images, videos of the real-world demonstration provide a more comprehensive understanding of our experiment. Our controller was successfully deployed on the Go1 quadruped robot for outdoor tests. As shown in Fig. \ref{fig::deployment1}, our robot exhibited excellent rhythm and beat synchronization with the music. Additionally, it demonstrated distinct gaits and periods when receiving different frequency commands, as shown in Fig. \ref{fig:fwalk}.

\section{CONCLUSIONS}

We investigated the phenomenon of natural rhythmic synchronization in animals and extended this concept to the gait actions of quadruped robots. The combination of a lower-level frequency tracker and oscillators forms a powerful gait generator, enables the robot to follow any nominal frequency and seamlessly transition into a stable trot gait. The higher-level phase modular ensures smooth movements and achieves precise synchronization with music. Experiments on the Go1 quadruped robot demonstrate real-time gait synchronization with music in real-world settings. This research improves the controllability of learning-based controllers through advanced rhythmic mechanisms, paving the way for real-time execution of complex rhythmic motions, particularly dancing, in robotic systems. Future research could explore synchronizing musical beats with foot trajectory planning to generate richer and more robust rhythmic motions.



\bibliographystyle{IEEEtran}
\bibliography{MW}

\addtolength{\textheight}{-12cm}   
\end{document}